\documentclass{article}

\usepackage{arxiv}

\usepackage[utf8]{inputenc} 
\usepackage[T1]{fontenc}    
\usepackage{hyperref}       
\usepackage{url}            
\usepackage{booktabs}       
\usepackage{amsfonts}       
\usepackage{nicefrac}       
\usepackage{microtype}      
\usepackage{lipsum}
\usepackage{graphicx}
\usepackage{subcaption}
\graphicspath{ {./images/} }
\usepackage{tabularx} 
\usepackage{amsmath}
\usepackage{float}
\usepackage[titletoc]{appendix}

\usepackage{soul}
\usepackage[colorinlistoftodos]{todonotes}

\title{CacheClip: Accelerating RAG with Effective KV Cache Reuse}

\author{
 Bin Yang \\
  Intel Corporation, Shanghai, China.\\
  \texttt{bin.yang@intel.com} \\
   \And
  Qiuyu Leng \\
  Intel Corporation, Shanghai, China. \\
  \texttt{qiuyu.leng@intel.com} \\
  \And
 Jun Zeng \\
  Intel Corporation, Chengdu, China. \\
  \texttt{jun1.zeng@intel.com} \\
  \And
  Zhenhua Wu \\
  Intel Corporation, Shanghai, China. \\
  \texttt{johnson.wu@intel.com} \\
}

\begin{document}
\maketitle
\begin{abstract}
Retrieval-Augmented Generation (RAG) systems suffer from severe time-to-first-token (TTFT) bottlenecks due to long input sequences. Existing KV cache reuse methods face a fundamental trade-off: prefix caching requires identical prefixes that rarely occur in RAG scenarios, while direct precomputation sacrifices quality due to missing inter-chunk attention and repeated attention sinks. Recent methods like APE and CacheBlend partially address these issues but remain inadequate for robust RAG applications.
This paper presents CacheClip, a novel framework that achieves both fast TTFT and high generation quality. 
Our key insight is that small auxiliary LLMs exhibit similar last-layer attention distributions to primary LLMs (the target model for generation), enabling efficient identification of tokens critical for restoring inter-chunk attention, thereby significantly improving response quality on cross-chunk reasoning tasks. 
CacheClip integrates four techniques: (1) auxiliary-model-guided token selection for selective KV cache recomputation, (2) shared prefixes to eliminate redundant attention sinks, (3) a sliding-window grouping strategy to maintain local coherence during partial KV cache updates, and (4) a CPU-GPU hybrid design that offloads auxiliary model inference to idle CPU resources, avoiding additional GPU overhead. The recomputation ratio is adjustable, allowing users to flexibly balance efficiency and quality for different deployment requirements.
Experiments show CacheClip retains up to 85.2\% and 91.1\% of full-attention performance on NIAH and LongBench, outperforming CacheBlend and APE by 16.1 and 12.8 points on NIAH, and by 4.5 and 4.2 points on LongBench (with recomp\% = 20\%). Meanwhile, CacheClip accelerates LLM inference by up to 3.33$\times$ in prefill time (with recomp\% = 20\%), providing a practical solution to the efficiency-quality trade-off in RAG systems.
\end{abstract}


\section{Introduction}



Large language models (LLMs) have demonstrated impressive capabilities in a wide range of applications~\cite{nazi2024large, li2023large, xia2025advancements, luo2025llm4sr}. 
However, their knowledge is inherently constrained by the training corpus: they are unable to capture up-to-date information and lack user-specific knowledge~\cite{cheng2024dated, kandpal2023large, li2024knowledge}.
To address these limitations and mitigate hallucinations~\cite{huang2025survey}, Retrieval-Augmented Generation (RAG) has emerged as a widely adopted paradigm~\cite{bechard2024reducing}. In RAG, a user query is prepended with multiple text chunks retrieved from an external database to provide domain knowledge or user-specific context~\cite{lewis2020retrieval}.
However, this paradigm comes with significant computational costs. RAG substantially increases the input length for large language models, which dramatically slows inference speed, particularly the time to first token (TTFT)~\cite{keles2023computational}. 
Given the wide application of RAG systems, addressing the TTFT bottleneck is a critical challenge.

A natural strategy is to precompute and reuse KV caches for retrieved text chunks, avoiding redundant computation across queries. However, existing reuse methods face fundamental trade-offs. Prefix caching~\cite{kwon2023efficient, zheng2023efficiently} requires identical input prefixes, a condition rarely met when retrieved chunks vary across queries. Direct KV cache concatenation~\cite{gim2024prompt} dramatically reduces TTFT but degrades quality due to missing inter-chunk attention and repeated attention sinks. Calibration methods such as APE~\cite{yang2025ape} address the attention sink problem but cannot recover cross-chunk dependencies, while CacheBlend's~\cite{yao2024cacheblend} selective recomputation relies on early-layer attention that poorly predicts which tokens matter in deeper layers, where the truly critical long-range dependencies are captured. Finetuning-based approaches~\cite{sun2024block, lu2024turborag, yang2025kvlink} incur high training costs and limited generalizability.

These limitations raise a key question: how can we accelerate RAG TTFT while preserving response quality?
In this paper, we propose CacheClip, a new framework designed to address both efficiency and quality challenges in RAG systems. Our main contributions are as follows:
\begin{itemize}
\item \textbf{Empirical observation:} We show that the last-layer attention distribution of small auxiliary LLMs is highly similar to that of primary LLMs (the target model for generation), which enables us to use a small auxiliary model to efficiently identify important tokens for recomputation in the primary LLM.
\item \textbf{Method:} We introduce CacheClip, which integrates four techniques: (1) auxiliary-LLM-guided selection of important tokens to restore inter-chunk dependencies, (2) a shared prefix to eliminate redundant attention sinks, (3) a sliding-window grouping strategy to maintain local coherence when partially updating the KV cache, and (4) a CPU-GPU hybrid design that offloads auxiliary model inference to idle CPU resources. The recomputation ratio is adjustable, enabling users to flexibly balance quality and efficiency based on deployment needs.
\item \textbf{Results:} We demonstrate that CacheClip retains up to 85.2\% and 91.1\% of full-attention performance on NIAH and LongBench, respectively, outperforming CacheBlend and APE by 16.1 and 12.8 points on NIAH, and by 4.5 and 4.2 points on LongBench (with recomp\% = 20\%). Meanwhile, CacheClip accelerates long-input RAG inference, achieving up to 3.33$\times$ speedup in prefill time.
\end{itemize}

\begin{figure}
  \centering
  \includegraphics[width=0.98\textwidth]{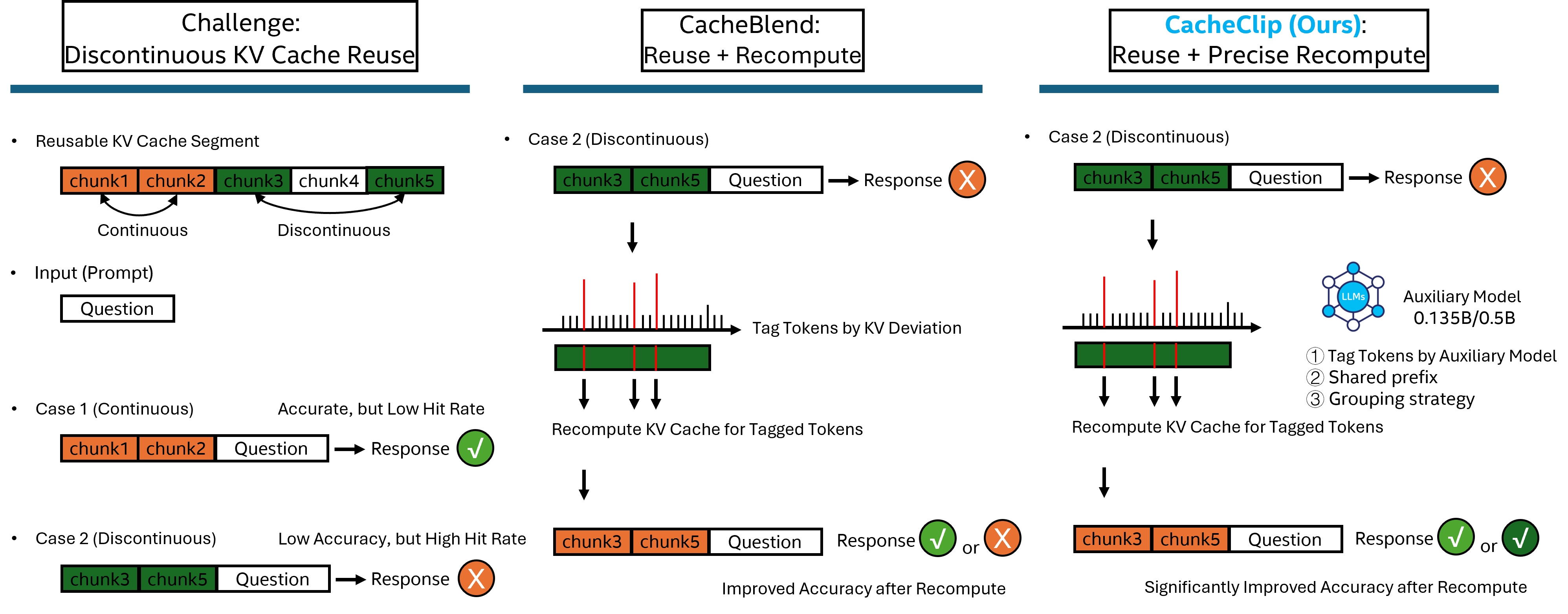}
  \caption{Illustration of CacheClip.}
  \label{fig:cacheclip_illustration}
\end{figure}

By effectively addressing the efficiency-quality trade-off, CacheClip significantly reduces inference latency in RAG systems while maintaining competitive generation quality. While we focus on RAG, the core idea of precomputing chunk-level KV caches and selectively recomputing critical tokens generalizes to other long-context scenarios such as agentic workflows where tools frequently inject large text blocks into the context. A high-level overview of CacheClip is illustrated in \autoref{fig:cacheclip_illustration}.

\section{Background and Motivation}
\label{sec:background}

\subsection{Prefill Latency in RAG Systems}
\label{subsec:prefill_latency}
In RAG systems, a user query is augmented with multiple text chunks retrieved from an external knowledge base, typically resulting in 4K to 16K input tokens per request~\cite{gao2023retrieval}. Processing these long inputs during the prefill stage requires attention computation that scales quadratically with the context length~\cite{vaswani2017attention}, leading to high time-to-first-token (TTFT) latency~\cite{shen2024towards}. Moreover, since RAG systems operate on a finite knowledge base, the same chunks are often retrieved across different queries~\cite{lewis2020retrieval, agarwal2025cache}. Fully recomputing their KV caches for every request wastes computation and limits system throughput under high concurrency.

A natural solution is to precompute and cache the KV caches for each text chunk, then reuse them at inference time by concatenating the caches of retrieved chunks. However, as we discuss below, existing approaches to KV cache reuse each face fundamental limitations.

\subsection{KV Cache Reuse: Approaches and Limitations}
\label{subsec:kv_reuse_limits}

\paragraph{Prefix caching.}
Prefix caching~\cite{kwon2023efficient, zheng2023efficiently} reuses the KV cache of previous requests when a new query shares an identical prefix. RAGCache~\cite{jin2024ragcache} extends this idea by constructing a prefix tree over retrieved chunk sequences. For example, if a previous request uses chunks \texttt{[D1, D2, D3]} and the next uses \texttt{[D1, D2, D5]}, only \texttt{[D1, D2]} can be reused. However, this reuse condition is overly strict: even minor changes in the query can alter the retrieved chunks and their order, limiting prefix caching to the first few chunks in practice. Furthermore, the same chunk appearing at different positions across queries requires storing multiple KV cache versions, incurring significant memory overhead.

\paragraph{Direct KV cache concatenation.}
A more direct approach precomputes each chunk's KV cache independently and concatenates them at inference time with adjusted positional embeddings~\cite{gim2024prompt}. While this dramatically reduces TTFT, it introduces two quality issues. First, computing each chunk in isolation eliminates cross-chunk attention, preventing the model from reasoning across multiple documents~\cite{yao2024cacheblend, yang2025kvlink}. Although the query tokens can attend to all chunk tokens during generation, this alone is insufficient to recover the rich inter-chunk dependencies that full attention provides. Second, transformers exhibit an attention sink effect~\cite{xiao2023efficient} where tokens at the start of each chunk receive disproportionately high attention. When independently computed caches are concatenated, these repeated attention sinks create patterns that diverge from training-time distributions, degrading generation quality.

\paragraph{KV cache calibration.}
To mitigate the repeated attention sink problem, APE~\cite{yang2025ape} and Zhang et al.~\cite{zhang2024attention} calibrate the concatenated KV cache by prepending a shared prefix to every chunk and retaining only the prefix from the first chunk. This removes redundant sink effects and improves consistency with training-time attention patterns. However, calibration alone cannot recover the missing inter-chunk attention. On tasks requiring multi-chunk reasoning, where answers must be composed from evidence dispersed across chunks, calibration-only methods still perform poorly (see \autoref{fig:ruler_recomp_combined}).

\paragraph{Finetuning the primary model.}
Block Attention~\cite{sun2024block}, TurboRAG~\cite{lu2024turborag}, and KVLink~\cite{yang2025kvlink} finetune LLMs to adapt to local attention mechanisms. However, finetuning a large primary model incurs high training costs, requires carefully curated datasets that balance chunk configurations and task diversity, and must be repeated whenever the base model or deployment setting changes.

\subsection{Selective Recomputation and the Token Selection Challenge}
\label{subsec:selective_recomp}

The analysis above reveals that effective KV cache reuse for RAG requires addressing two issues simultaneously: (1) eliminating repeated attention sinks, and (2) restoring the missing inter-chunk attention. Calibration handles the first issue. For the second, \textit{selective recomputation}, i.e., updating the KV cache for only a small subset of important tokens, offers a promising direction.

CacheBlend~\cite{yao2024cacheblend} pioneered this approach by selecting tokens to recompute based on the primary model's early-layer attention. While this demonstrates the feasibility of selective recomputation, its effectiveness relies on accurate token selection: choosing the wrong tokens wastes the recomputation budget without recovering the critical inter-chunk dependencies.

As we will show in Section~\ref{sec:analysis_observations}, early-layer signals are fundamentally limited for this purpose, as shallow layers capture local syntactic patterns rather than the long-range semantic dependencies that matter in deep layers. This motivates the central question of our work:

\vspace{0.3em}
\noindent\textit{Given precomputed local KV caches for retrieved text chunks, how can we accurately identify the most important tokens for selective recomputation, so as to restore inter-chunk attention and maximize generation quality within a limited recomputation budget?}
\vspace{0.3em}

\noindent In the next section, we present the key observation that answers this question: small auxiliary LLMs can identify these critical tokens far more accurately than the primary model's own early layers.

\section{Analysis and Key Observations}
\label{sec:analysis_observations}

As discussed in Section~\ref{sec:background}, selective recomputation offers a promising pathway to restore inter-chunk attention without the prohibitive cost of full recomputation. However, the core bottleneck lies in accurately identifying which tokens are most critical for recomputation during the prefill stage. In this section, we analyze this problem from three angles: (1) the inherent sparsity of attention that makes selective recomputation feasible, (2) the fundamental challenge of token selection during prefill and why early-layer signals are inadequate, and (3) our key finding that small auxiliary models can effectively capture the deep-layer attention patterns of much larger models.

\subsection{Attention Sparsity Enables Selective Recomputation}
\label{subsec:sparsity}

The feasibility of restoring cross-chunk attention through selective recomputation relies on a well-established property of Transformer attention: \textit{sparsity}.

Extensive research has demonstrated that attention distributions in large language models are highly sparse~\cite{chen2021scatterbrain, zhang2023h2o}. Rather than attending uniformly to all tokens, the model concentrates most of its attention weight on a small subset of contextually important tokens. Empirically, the top 10 to 20\% of tokens often account for the majority of the total attention weight~\cite{zhang2023h2o, li2024snapkv}. This concentration is consistent across model scales and task types, reflecting the model's learned ability to focus on semantically critical information.

This sparsity provides the theoretical foundation for selective recomputation: if we can accurately identify the most important tokens, we can recover the essential inter-chunk attention paths by recomputing only their KV caches. The remaining tokens, which receive minimal attention weight, contribute little to the final output and can safely retain their precomputed local KV values.

However, attention sparsity patterns are highly \textit{dynamic} and query-aware~\cite{jiang2024minference, li2024snapkv}. The importance of a token shifts dramatically depending on the specific user query and surrounding context. This means that important tokens cannot be identified statically during offline preprocessing: token selection must be performed at runtime, tailored to the current input.

\subsection{The Challenge of Token Selection During Prefill}
\label{subsec:challenge}

Given that attention sparsity makes selective recomputation feasible, the critical question becomes: \textit{how can we accurately identify which tokens to recompute?}

\paragraph{Unique difficulties during prefill.}
Unlike the decoding stage where the full KV cache of all previous tokens is available, the prefill stage of a KV-cache-reuse system only has \textit{local} KV caches for individual chunks, lacking inter-chunk attention. This makes it infeasible to compute exact attention scores with respect to the complete concatenated context. An effective token selection method must therefore be both query-aware and computationally lightweight, operating without full-context attention.

\paragraph{CacheBlend's early-layer approach.}
CacheBlend~\cite{yao2024cacheblend} addresses this by using signals from the primary model's early layers. It fully recomputes the first layer and partially recomputes the second layer, then compares the resulting value matrix (V) against the precomputed version to identify tokens with the largest discrepancies.

\paragraph{Why early layers are insufficient.}
However, this early-layer-based selection is fundamentally limited. Prior studies~\cite{jawahar2019does, vig2019analyzing, cai2024pyramidkv} have established that different Transformer layers serve distinct semantic functions: shallow layers primarily capture local syntactic relationships, while deeper layers handle long-range semantic dependencies and task-oriented reasoning. Consequently, the set of tokens that receive high attention in early layers differs substantially from the set that matters in deeper layers. Using shallow-layer signals to predict deep-layer token importance introduces a systematic bias that cannot be resolved by simply increasing the recomputation budget.

This layer-wise functional divergence will be quantitatively verified in Section~\ref{subsec:small_but_aligned}, where we show that the primary model's own first-layer attention has low overlap with its last-layer attention (as measured by Jaccard Index of top-$k$ tokens). This observation leads to a natural question: if the primary model's own early layers cannot predict its deep-layer behavior, can we instead use an \textit{external lightweight model's deep layers} to approximate the primary model's deep-layer attention?

\subsection{Small Models Capture Large Models' Attention Patterns}
\label{subsec:small_but_aligned}

We now investigate whether a small auxiliary model can approximate the attention patterns of a much larger primary model, thereby guiding token selection more effectively than the primary model's own early layers.

\paragraph{Experimental setup.}
We evaluate four model pairs spanning both same-family and cross-family configurations, as shown in \autoref{tab:model_pairs}. We use 200 samples from 2WikiMultihopQA~\cite{xanh2020_2wikimultihop} at each of five sequence lengths (1K, 2K, 4K, 8K, 16K tokens). For each sample, we extract the head-averaged last-token attention distribution from both the first and last layers of each model, and compare them using the following protocol:
\begin{itemize}
    \item \textbf{Aux vs.\ Primary} (\textit{aux\_last} $\leftrightarrow$ \textit{pri\_last}): the auxiliary model's last-layer attention vs.\ the primary model's last-layer attention.
    \item \textbf{Primary Internal} (\textit{pri\_first} $\leftrightarrow$ \textit{pri\_last}): the primary model's own first-layer attention vs.\ its last-layer attention.
\end{itemize}
Since different models use different tokenizers (e.g., SmolLM2-135M produces ${\sim}1050$ tokens for the same text where Qwen2.5-14B produces ${\sim}1024$), direct token-level comparison is unfair. We therefore project all attention distributions into a unified \textit{character space}: using each tokenizer's offset mapping, we uniformly distribute each token's attention weight across the character span it covers. All metrics are then computed in this shared character space.

Our primary metric is the \textbf{Jaccard Index} of the top-20\% highest-attention character positions, which directly measures the overlap of ``important positions'' between two distributions, which corresponds precisely to CacheClip's token selection task. We additionally report \textbf{KL divergence} as a supplementary metric capturing overall distributional similarity.

\begin{table}[t]
\centering
\caption{Model pairs used in the attention alignment study. ``Cross-family'' pairs use models from different architectures with different tokenizers.}
\label{tab:model_pairs}
\small
\begin{tabular}{@{}llll@{}}
\toprule
Pair & Auxiliary Model & Primary Model & Relation \\
\midrule
A & Qwen2.5-0.5B-Instruct & Qwen2.5-14B-Instruct & Same family \\
B & SmolLM2-135M-Instruct & Qwen2.5-14B-Instruct & Cross-family \\
C & SmolLM2-135M-Instruct & LLaMA-3.1-8B-Instruct & Cross-family \\
D & SmolLM2-135M-Instruct & Ministral-8B-Instruct & Cross-family \\
\bottomrule
\end{tabular}
\end{table}

\begin{figure}[t]
  \centering
  \includegraphics[width=0.98\textwidth]{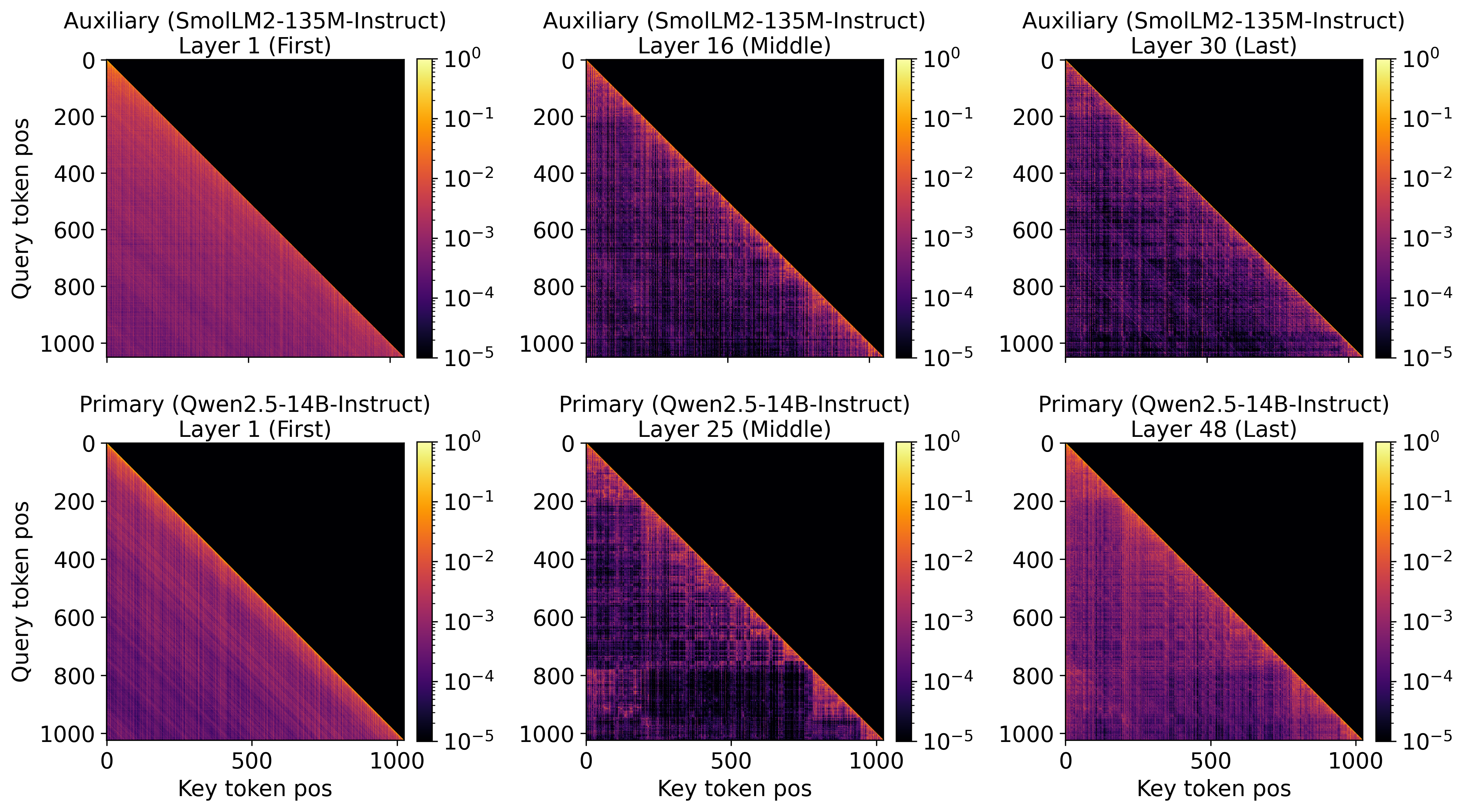}
  \caption{Attention heatmaps (head-averaged, log scale) at Layer 1, middle layer, and last layer for SmolLM2-135M (top row) and Qwen2.5-14B (bottom row) on the same 1K-token input. The last layers of both models (c, f) show similar sparse attention patterns, while the primary model's first layer (d) is notably more diffuse.}
  \label{fig:attention_heatmap}
\end{figure}

\paragraph{Qualitative evidence: attention heatmaps.}
Before presenting aggregate statistics, we provide visual evidence of cross-model attention alignment. \autoref{fig:attention_heatmap} shows the full attention matrix (averaged over heads) at three representative layers (first, middle, and last) for SmolLM2-135M (auxiliary) and Qwen2.5-14B (primary) on the same input. Despite the 100$\times$ difference in model size and entirely different architectures, the last-layer attention patterns (panels (c) and (f)) exhibit strikingly similar sparse structures, with attention concentrated on the same token positions. In contrast, the primary model's first layer (panel (d)) shows a diffuse, qualitatively different pattern from its own last layer (panel (f)), visually confirming the layer-wise functional divergence discussed in Section~\ref{subsec:challenge}.

\begin{figure}[t]
  \centering
  \includegraphics[width=0.55\textwidth]{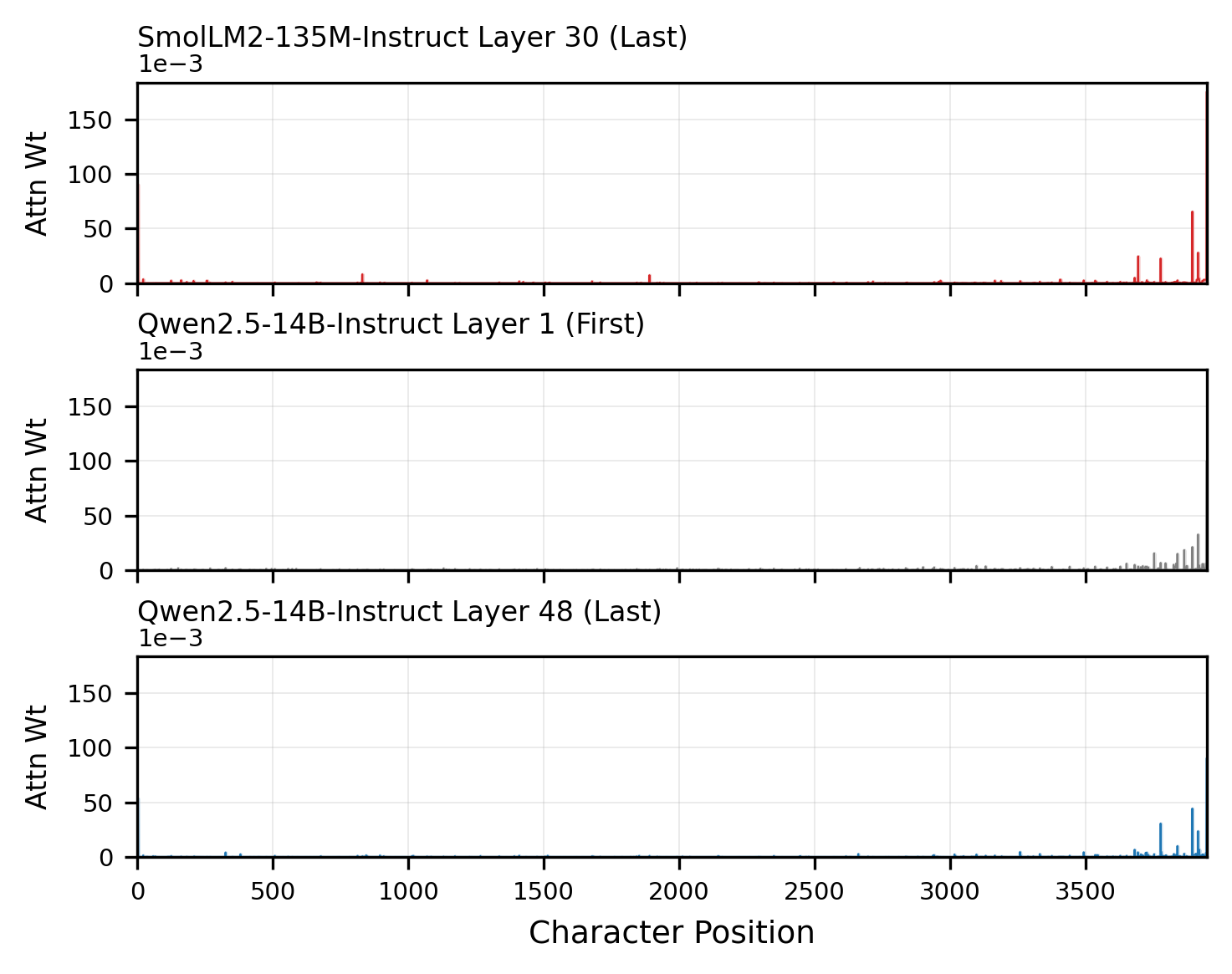}
  \caption{Last token's attention over context in character space. Top: SmolLM2-135M-Instruct last layer. Middle: Qwen2.5-14B-Instruct first layer. Bottom: Qwen2.5-14B-Instruct last layer. (Pearson/Spearman: aux$_\text{last}$ vs.\ pri$_\text{last}$: $0.97/0.60$; pri$_\text{first}$ vs.\ pri$_\text{last}$: $0.70/0.54$.)}
  \label{fig:attention_1d_curve}
\end{figure}

\autoref{fig:attention_1d_curve} further illustrates this alignment by plotting the last token's attention distribution over all preceding positions, projected into character space. The auxiliary model's last layer and the primary model's last layer show highly correlated peaks (Pearson $r = 0.97$, Spearman $\rho = 0.60$), while the primary model's own first layer is less aligned with its last layer (Pearson $r = 0.70$, Spearman $\rho = 0.54$).

\paragraph{Core result: small models identify the right token positions.}
\autoref{tab:jaccard_results} presents the Jaccard Index of top-20\% attention positions across all four model pairs and four sequence lengths, averaged over 200 samples each. The central finding is clear: \textbf{in all 16 configurations, the auxiliary model's last layer achieves higher top-$k$ position overlap with the primary model's last layer than the primary model's own first layer does.} In other words, even a 135M-parameter model from a different architecture family identifies the primary model's important token positions more accurately than the primary model's own early layers.

This result is particularly noteworthy for the cross-family pairs (B, C, D), where the auxiliary and primary models have different architectures, different training data, and different tokenizers. The consistency of this finding across all pairs demonstrates that the phenomenon is not an artifact of shared model families; it reflects a more fundamental property of how Transformer models converge on similar notions of token importance in their deep layers.

\begin{table}[t]
\centering
\caption{Jaccard Index (top-20\% character positions) between attention distributions. Each cell shows \textit{aux\_last vs.\ pri\_last} / \textit{pri\_first vs.\ pri\_last}, averaged over 200 samples. Higher is better for aux vs.\ pri; the key observation is that aux vs.\ pri consistently exceeds pri internal across all settings.}
\label{tab:jaccard_results}
\small
\begin{tabular}{@{}lccccc@{}}
\toprule
Model Pair & 1K & 2K & 4K & 8K & 16K \\
\midrule
A: Qwen0.5B $\to$ Qwen14B   & \textbf{0.41} / 0.23 & \textbf{0.41} / 0.22 & \textbf{0.39} / 0.22 & \textbf{0.39} / 0.22 & \textbf{0.39} / 0.22 \\
B: SmolLM $\to$ Qwen14B      & \textbf{0.30} / 0.23 & \textbf{0.29} / 0.22 & \textbf{0.29} / 0.22 & \textbf{0.28} / 0.22 & \textbf{0.31} / 0.22 \\
C: SmolLM $\to$ LLaMA8B      & \textbf{0.31} / 0.13 & \textbf{0.31} / 0.13 & \textbf{0.29} / 0.12 & \textbf{0.28} / 0.12 & \textbf{0.28} / 0.11 \\
D: SmolLM $\to$ Ministral8B  & \textbf{0.32} / 0.19 & \textbf{0.32} / 0.17 & \textbf{0.31} / 0.17 & \textbf{0.29} / 0.16 & \textbf{0.34} / 0.16 \\
\bottomrule
\end{tabular}
\end{table}

\paragraph{Supplementary metric: KL divergence confirms distributional alignment within families.}
As a supplementary analysis, we also measure KL divergence between the same pairs of attention distributions (\autoref{tab:kl_results}). For pairs where the primary model is Qwen2.5-14B (Pairs A and B), $\text{KL}(\textit{aux\_last} \| \textit{pri\_last})$ is consistently lower than $\text{KL}(\textit{pri\_first} \| \textit{pri\_last})$, confirming that the auxiliary model's attention distribution is closer in shape to the primary model's last layer than the primary model's own first layer is. For cross-family pairs with different primary models (Pairs C and D), the KL relationship reverses. This is expected, as different architectures naturally produce attention distributions with different ``stylistic'' properties (e.g., varying degrees of peakiness), which inflates KL divergence. Crucially, this does not affect the token \textit{position} selection task: as shown by Jaccard, the \textit{locations} of high-attention tokens remain well-aligned even when the distribution \textit{shapes} differ across architectures. Since CacheClip's token selection only requires identifying the right positions, not replicating the exact distribution, Jaccard is the metric that directly matters.

\begin{table}[t]
\centering
\caption{KL divergence between attention distributions (in character space). Each cell shows \textit{aux\_last vs.\ pri\_last} / \textit{pri\_first vs.\ pri\_last}. Lower aux vs.\ pri indicates better alignment. KL confirms distributional alignment for same-primary pairs (A, B); cross-family pairs show higher KL due to architectural style differences, which does not affect position-level agreement (see Jaccard in \autoref{tab:jaccard_results}).}
\label{tab:kl_results}
\small
\begin{tabular}{@{}lccccc@{}}
\toprule
Model Pair & 1K & 2K & 4K & 8K & 16K \\
\midrule
A: Qwen0.5B $\to$ Qwen14B   & \textbf{0.58} / 0.92 & \textbf{0.57} / 0.94 & \textbf{0.64} / 1.11 & \textbf{0.62} / 1.13 & \textbf{0.64} / 1.12 \\
B: SmolLM $\to$ Qwen14B      & \textbf{0.33} / 0.92 & \textbf{0.37} / 0.94 & \textbf{0.46} / 1.11 & \textbf{0.50} / 1.13 & \textbf{0.67} / 1.12 \\
C: SmolLM $\to$ LLaMA8B      & 0.74 / \textbf{0.28} & 0.72 / \textbf{0.31} & 0.76 / \textbf{0.33} & 0.76 / \textbf{0.36} & 1.06 / \textbf{0.38} \\
D: SmolLM $\to$ Ministral8B  & 1.32 / \textbf{0.80} & 1.40 / \textbf{0.93} & 1.64 / \textbf{1.13} & 1.80 / \textbf{1.30} & 1.68 / \textbf{1.54} \\
\bottomrule
\end{tabular}
\end{table}

\paragraph{Summary.}
Our experiments demonstrate that even a 135M-parameter auxiliary model can accurately identify the token positions that 8B to 14B primary models attend to most strongly in their final layers, and does so more reliably than the primary model's own early layers. This finding is consistent across four model pairs, five sequence lengths, and 200 samples per configuration, encompassing both same-family and cross-architecture settings. It provides the empirical foundation for CacheClip's design: using the auxiliary model's last-layer attention to guide token selection during prefill, replacing the unreliable early-layer approach of prior methods such as CacheBlend.

\section{Methodology of CacheClip}

Motivated by the key observations in Section~\ref{sec:analysis_observations}, we propose CacheClip, a novel framework designed to accelerate RAG inference while preserving generation quality. The core philosophy of CacheClip is to reuse the majority of precomputed local KV caches for efficiency, while leveraging a lightweight auxiliary model to dynamically identify and recompute only the most critical tokens to restore inter-chunk attention.

\begin{figure}
  \centering
  \includegraphics[width=0.98\textwidth]{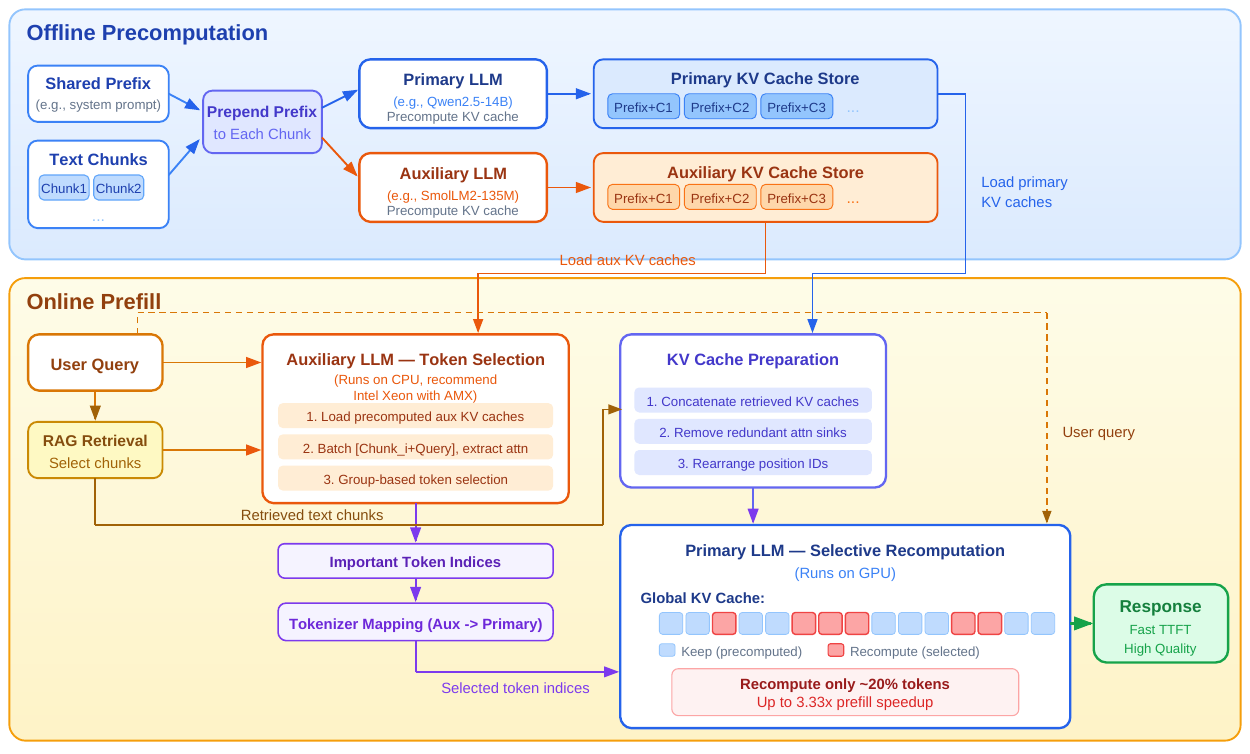}
  \caption{Workflow of CacheClip.}
  \label{fig:cacheclip_workflow}
\end{figure}

As illustrated in \autoref{fig:cacheclip_workflow}, the workflow of CacheClip consists of two phases. In the \textbf{offline phase}, we precompute local KV caches for each text chunk with a shared prefix prepended (Section~\ref{subsec:attention_sink}). In the \textbf{online phase}, upon receiving a user query, we concatenate the precomputed chunk KV caches, rearrange position IDs, and retain only one copy of the shared prefix to calibrate attention distributions. Then, an auxiliary LLM identifies important tokens for recomputation based on the user query (Section~\ref{subsec:token_selection}), a sliding-window grouping strategy organizes these tokens into contiguous groups to preserve contextual integrity, and the primary LLM selectively recomputes and updates the global KV cache (Section~\ref{subsec:grouping}). The auxiliary model runs on CPU to avoid consuming GPU resources (Section~\ref{subsec:hybrid}).

\subsection{KV Cache Calibration with Shared Prefix}
\label{subsec:attention_sink}
To mitigate the repeated attention sink effects~\cite{xiao2023efficient} that arise at the beginning of each independently processed text chunk, CacheClip adopts a shared prefix strategy during KV cache precomputation. Specifically, inspired by~\cite{yang2025ape, zhang2024attention}, we prepend a fixed prefix to each text chunk before feeding it into the LLM. In our implementation, we use the system prompt as the shared prefix by default.

Once relevant text chunks are retrieved by the RAG system, we concatenate their KV caches and eliminate redundant attention sinks by keeping only the shared prefix from the first chunk. As a result, the final KV cache contains only one attention sink (at the beginning of the full input), thereby restoring the global attention distribution to be more consistent with the patterns observed during LLM training. 

This calibration step addresses the attention distribution misalignment and contributes to more coherent and accurate model outputs.


\paragraph{Position ID rearrangement.}
After removing redundant shared prefixes, the position IDs across chunks exhibit unintended repetition, since each chunk was originally precomputed starting from the same offset following the shared prefix. CacheClip rearranges the position IDs so that each chunk is assigned a unique and continuous position range following the preceding chunk, forming a monotonically increasing sequence consistent with full attention mode.

\subsection{Auxiliary-Model-Guided Token Selection}
\label{subsec:token_selection}
Instead of comparing value matrices from the first layer~\cite{yao2024cacheblend}, we propose using a auxiliary LLM to identify important tokens, which are then recomputed in the primary LLM (the target model for generation).

As discussed in \autoref{subsec:small_but_aligned}, the attention patterns in the final layers of small auxiliary LLMs are generally aligned with those of primary LLMs. While auxiliary LLMs lack the capacity to answer complex queries over long documents, they still effectively capture which tokens are most relevant to the query. In other words, tokens with high attention scores in the auxiliary model are often also important to the primary model.

To improve efficiency, we also precompute the KV cache for each retrieved text chunk in the auxiliary LLM. Then, we concatenate each chunk with the user query and process them in batch (i.e., \texttt{chunk1 + query}, \texttt{chunk2 + query}, ...). This allows us to reuse the precomputed chunk-side KV cache, significantly reducing computation during token selection.

We extract the attention matrix from the final layer of the auxiliary LLM. To focus on query-relevant information, we isolate the attention weights from query tokens to chunk tokens, excluding attention among query tokens and shared prefix tokens. The resulting matrix has shape $\mathopen{[}\text{query\_size}, \text{total\_chunk\_size}\mathclose{]}$. By averaging over the query dimension, we obtain a single importance score per token, resulting in a final vector of shape $\mathopen{[}\text{total\_chunk\_size}\mathclose{]}$.

Based on these importance scores, we select the top-$k$ tokens (where $k$ is determined by the recomputation ratio) as candidate tokens for recomputation. These candidates are then passed to the sliding-window grouping stage (Section~\ref{subsec:grouping}) to form contiguous groups for recomputation.

\subsection{Sliding-Window Grouping and KV Cache Update}
\label{subsec:grouping}

Recomputing only sparse, scattered tokens can break local contextual coherence: when some tokens within a semantically coherent group are updated while the rest retain stale values, the resulting KV cache becomes internally inconsistent. To address this, we adopt a sliding-window grouping strategy that organizes candidate tokens into contiguous groups for recomputation.

Given the set of candidate tokens selected by the auxiliary model, we scan the sequence using a sliding window of size $w$ (default $w{=}8$) with a step size of 1. A window is evaluated only when its starting position is a candidate token. Within each such window, we count the number of candidate tokens (i.e., the local density). If the count meets or exceeds a predefined threshold $\tau$ (default $\tau{=}5$), we include \emph{all} tokens in that window, including non-candidate tokens, into the final recomputation set. This ensures that recomputed tokens form contiguous groups, preserving local contextual integrity.

Candidate tokens that are not covered by any window meeting the density threshold are considered isolated and are dropped from the recomputation set, avoiding fragmentary updates that would harm coherence.

\paragraph{Token mapping and KV cache update.}
Since the auxiliary and primary LLMs may use different tokenizers, we perform token-level alignment to map the selected token indices from the auxiliary tokenizer to the primary tokenizer. The primary LLM then recomputes the KV cache for these tokens with position IDs consistent with full attention mode, and selectively overwrites the corresponding entries in the global KV cache.




\subsection{CPU-GPU Hybrid System Design}
\label{subsec:hybrid}
Although CacheClip introduces an auxiliary LLM for token selection, it does not require extra GPU memory or GPU compute, which avoids workload imbalance across multiple GPU cards.
Instead, the auxiliary LLM runs on the head node's CPU, whose resources are typically idle during GPU-based inference of the primary LLM.
With the auxiliary LLM's KV cache precomputation, the token selection workload is lightweight in terms of FLOPs. This workload can be handled efficiently by the CPU. An Intel Xeon CPU equipped with an AMX accelerator~\cite{intelamx} can further accelerate the matrix operations involved in token selection~\cite{abouelhamayed2025sparamx}.

In addition, the token selection process can be overlapped with the KV cache loading of the primary LLM, hiding most of its latency. The recomputation ratio can also be dynamically adjusted at runtime according to the accuracy and latency requirements of the task.

Under this hybrid design, CacheClip avoids consuming additional GPU resources while effectively reducing the first-token latency.

\section{Experiments}
In this section, we present the effectiveness and efficiency of CacheClip. 
Specifically, we aim to answer the following questions: 
1) How does CacheClip perform compared to baseline methods? 
2) Can auxiliary models effectively guide token selection during prefill? 
3) How does CacheClip perform under different recomputation budgets and sequence lengths?
We empirically investigate these questions through the following experiments.

\subsection{Setup}

\paragraph{Models and hardware settings}
We evaluate CacheClip on Qwen2.5-14B-Instruct~\cite{qwen2.5} as the primary model, with SmolLM2-135M-Instruct~\cite{allal2025smollm2smolgoesbig} and Qwen2.5-0.5B-Instruct as auxiliary models.
Our end-to-end experiments are conducted on NVIDIA L20 GPUs, with the head node equipped with 5th Gen Intel Xeon EMR CPUs 6554S.

\paragraph{Dataset \& Evaluation Metrics}  
We use the provided metrics and scripts from the following benchmarks for evaluation.
\begin{itemize}
\item RULER~\cite{hsieh2024ruler}: We use its retrieval category, which extends needle-in-a-haystack (NIAH)~\cite{kamradt2023needle} into eight variations with diverse types and quantities of ``needles'' and ``haystacks,'' evaluating a model's ability to retrieve and aggregate specific information from long contexts.
We test with sequence lengths ranging from 4K to 16K tokens to cover typical RAG input lengths.
Following RULER, we adopt the average substring match rate as the evaluation metric, referred to as Average Reference Coverage (ARC).
\item LongBench~\cite{bai2023longbench}: We use the multifieldqa\_zh, 2wikimqa, and hotpotqa datasets to evaluate long-context understanding across multiple tasks. 
Following LongBench, we utilize automatic metrics such as ROUGE-L and F1 to measure similarity to ground-truth answers.
\end{itemize}
To evaluate LLM long-input processing capability without being affected by retriever quality, we bypass the retriever stage and directly use all text chunks. Specifically, we split the input into chunks of 1000 tokens with 50-token overlaps, precompute the KV cache for each chunk independently, and concatenate them as input to CacheClip and the baseline methods that support KV cache reuse.

\paragraph{Baseline} We compare CacheClip with the following baselines:
\begin{itemize}
\item Full Attention: The raw contexts are fed into the LLM, which computes KV caches for all tokens during prefill.
\item Direct reuse: Directly concatenate precomputed KV caches with positional embeddings adjusted accordingly.
\item APE~\cite{yang2025ape}: Calibrates KV cache via shared prefix and attention rescaling (temperature 0.9, scaling factor 0.9).
\item CacheBlend~\cite{yao2024cacheblend}: Recomputes a subset of tokens to selectively update the KV cache. It identifies top-$k$ differing tokens using the value matrix (V) from the second layer.
\end{itemize}

\subsection{RULER Analysis}

\begin{figure}
  \centering
  \includegraphics[width=0.85\textwidth]{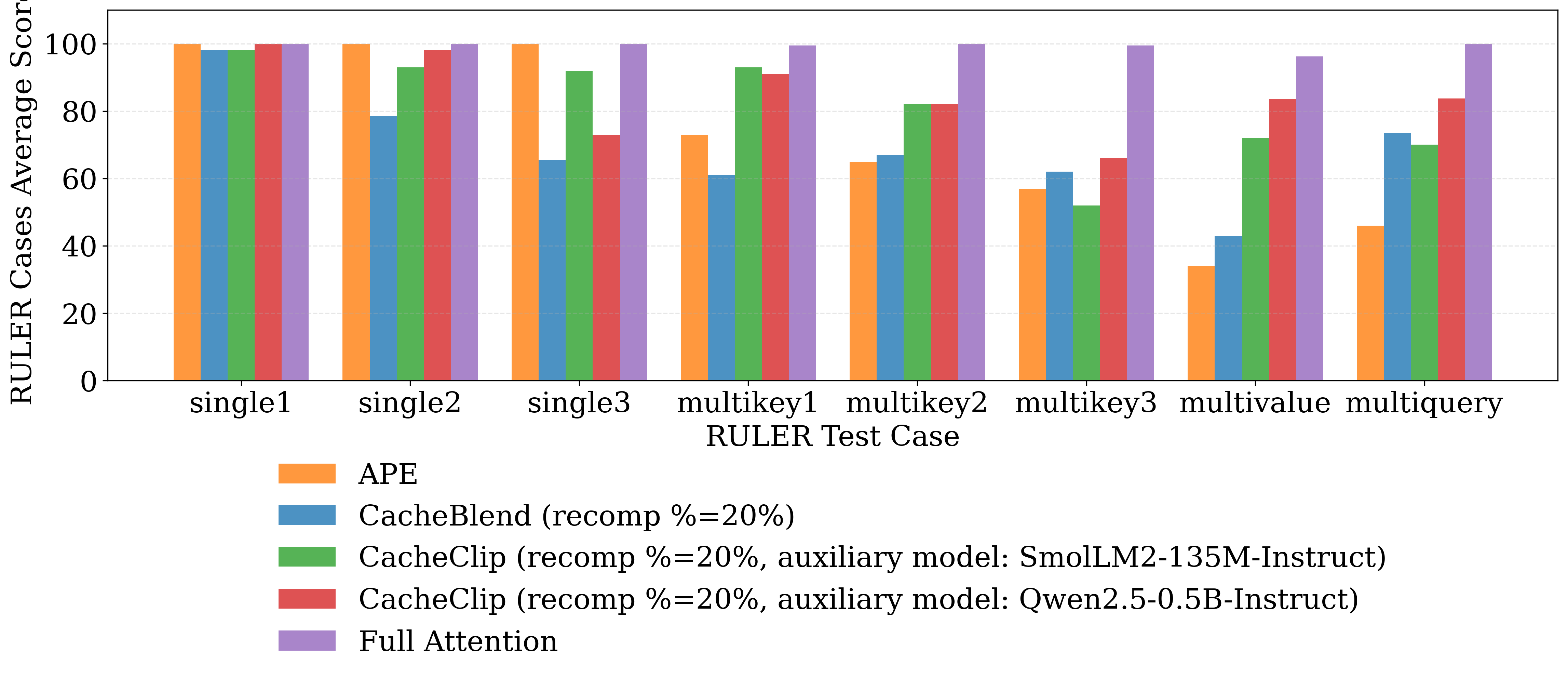}
  \caption{Performance across RULER test cases (Qwen2.5-14B-Instruct, input length = 8192, recomp\% = 20\%).}
  \label{fig:ruler_case_comparison}
\end{figure}

As shown in \autoref{fig:ruler_case_comparison}, CacheClip achieves strong performance across all RULER test cases with only 20\% recomputation. The multivalue case is particularly representative of real-world RAG scenarios, requiring the model to identify and aggregate four values scattered across different chunks. CacheClip achieves 96.00 on this task, while APE scores only 34.00 (no recomputation to recover cross-chunk attention) and CacheBlend scores 42.97 (early-layer selection fails to prioritize cross-chunk tokens).

CacheBlend also exhibits an unexpected failure mode on single2/single3: as shown in \autoref{tab:cacheblend_ruler}, its performance drops sharply as the recomputation ratio increases from 5\% to 80\%, a counterintuitive result where more recomputation leads to worse quality. The target values (7-digit numbers, UUIDs) are split into multiple tokens by the tokenizer. When CacheBlend recomputes only part of the value tokens, the fused KV cache breaks context continuity, producing corrupted outputs. CacheClip avoids this through sliding-window grouping (\autoref{subsec:grouping}), which ensures nearby tokens are recomputed together as contiguous groups.

\subsection{Impact of Recomputation Ratio and Sequence Length}
\label{subsec:recomp_ratio}

\begin{figure}[htbp]
  \centering
  \includegraphics[width=\textwidth]{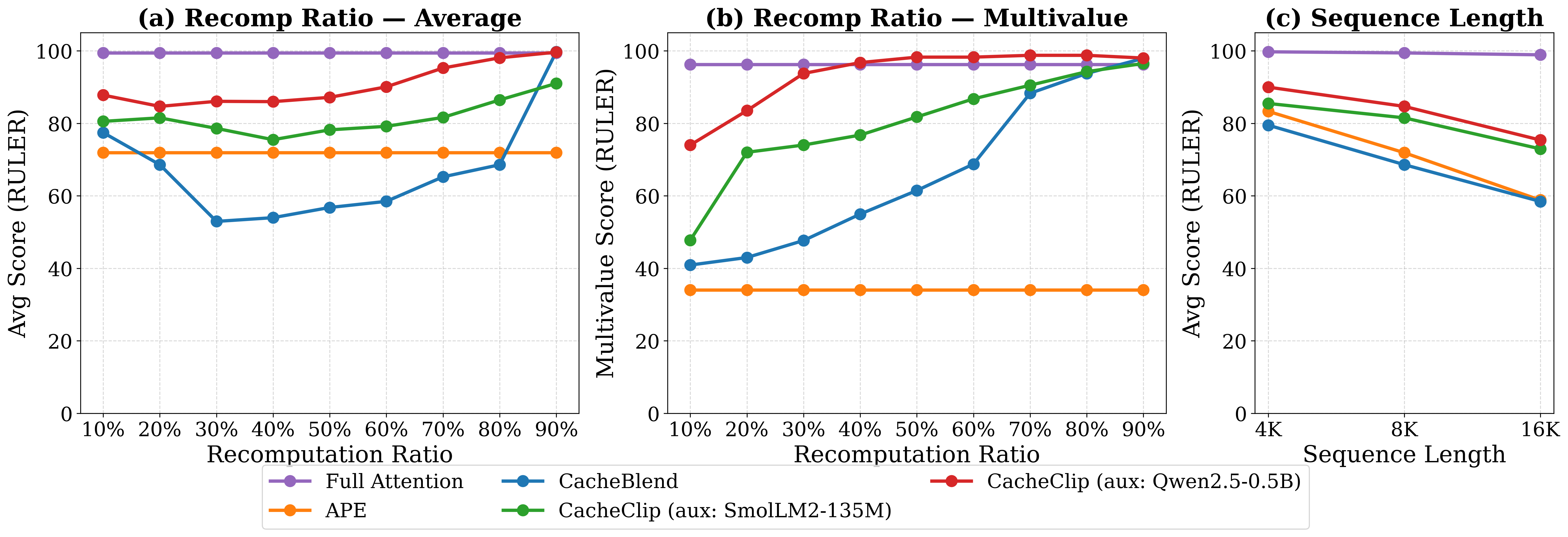}
  \caption{Left \& Middle: Impact of recomputation ratio on RULER (Qwen2.5-14B-Instruct, input length = 8192). Right: Impact of sequence length on RULER (recomp\% = 20\%).}
  \label{fig:ruler_recomp_combined}
\end{figure}

\autoref{fig:ruler_recomp_combined} (left, middle) shows the recomputation ratio sweep. CacheClip's curve rises steeply at low ratios and saturates quickly: with 10\% recomputation it already scores $\sim$90, reaching over 94 at 20\%. This rapid saturation confirms that the auxiliary model selects the most critical tokens first, consistent with the attention sparsity discussed in Section~\ref{subsec:sparsity}.

In contrast, CacheBlend exhibits a U-shaped average curve (declining in the 30 to 80\% range before recovering) and an S-shaped multivalue curve. On multivalue, CacheClip achieves 96\% at 20\% recomputation while CacheBlend reaches only 43\%, not surpassing 90\% until 70\% recomputation. This means CacheClip recovers the same cross-chunk attention with $3.5\times$ fewer recomputed tokens.

\autoref{fig:ruler_recomp_combined} (right) evaluates all methods at 4K, 8K, and 16K tokens with 20\% recomputation. From 4K to 16K, APE drops by 24.5 points, CacheBlend by 21.0, while CacheClip drops by only 12.5 (SmolLM2-135M) and 14.6 (Qwen2.5-0.5B) points, demonstrating that auxiliary-model-guided selection scales better with context length. At 16K, the baselines converge to $\sim$58 while CacheClip maintains 72.9 and 75.4 with the two auxiliary models.

\subsection{LongBench Analysis}

\begin{figure}
  \centering
  \includegraphics[width=0.80\textwidth]{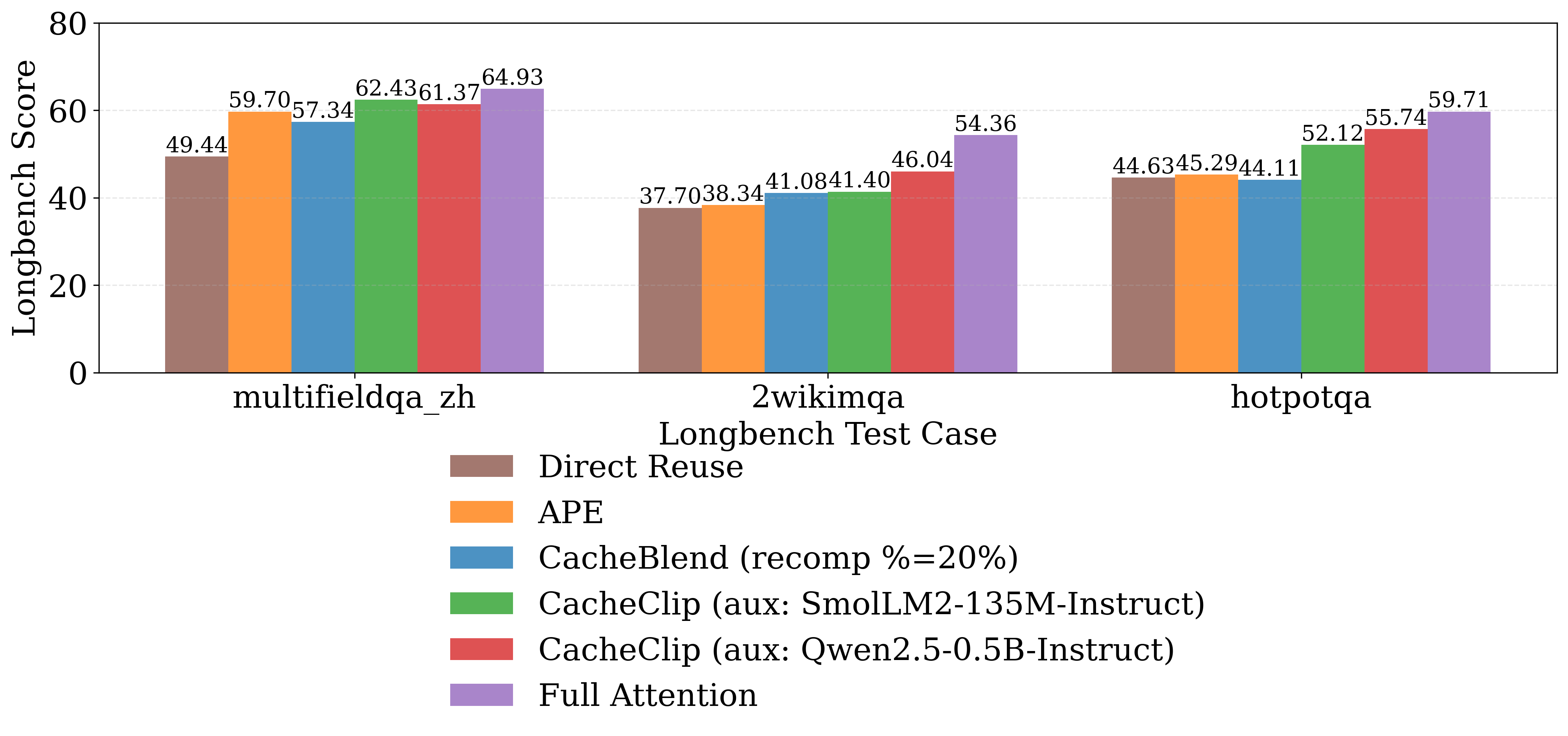}
  \caption{Performance across LongBench test cases (Qwen2.5-14B-Instruct, recomp\% = 20\%).}
  \label{fig:longbench_case_comparison}
\end{figure}

To validate beyond synthetic tasks, we evaluate on LongBench (\autoref{fig:longbench_case_comparison} and \autoref{tab:longbench}). CacheClip consistently outperforms all KV cache reuse baselines across three datasets. On hotpotqa (multi-hop QA requiring cross-chunk reasoning), CacheClip (SmolLM2-135M) exceeds CacheBlend by +8.0 and APE by +6.8 points. With the stronger Qwen2.5-0.5B auxiliary model, performance further improves to 91.1\% of full-attention quality. Detailed per-dataset results are in \autoref{tab:longbench}.

\subsection{Ablation Study}
\label{subsec:ablation}

We ablate CacheClip's components on RULER (\autoref{table:ablation_ruler}) and LongBench (\autoref{table:ablation_longbench}).

\begin{table}[h]
 \caption{Ablation on RULER (Qwen2.5-14B-Instruct, seq len = 8192, recomp\% = 20\%).}
  \centering
  \resizebox{\textwidth}{!}{%
  \begin{tabular}{cc@{\hskip 1.2em}rrrrrrrr@{\hskip 1.2em}r}
    \toprule
    \multicolumn{2}{c}{Components} & \multicolumn{9}{c}{RULER Scores} \\
    \cmidrule(lr){1-2} \cmidrule(lr){3-11}
    Shared Prefix & Grouping & single1 & single2 & single3 & multikey1 & multikey2 & multikey3 & multivalue & multiquery & Avg. \\
    \midrule
    \multicolumn{2}{c}{\textit{Full Attention (reference)}} & 100 & 100 & 100 & 99.50 & 100 & 99.50 & 96.25 & 100 & 99.41 \\
    \midrule
                 &              & 78.00  & 20.00  & 39.00  & 33.00  & 77.00  & 20.00  & 25.25  & 31.50  & 40.47  \\
    \checkmark   &              & 64.00  & 20.00  & 46.00  & 31.00  & 74.00  & 16.00  & 26.25  & 32.00  & 38.66  \\
                 & \checkmark   & \textbf{98.00}  & 92.00  & 81.00  & \textbf{94.00}  & \textbf{86.00}  & \textbf{55.00}  & \textbf{74.50}  & 69.75  & 81.28  \\
    \checkmark   & \checkmark   & \textbf{98.00}  & \textbf{93.00}  & \textbf{92.00}  & 93.00  & 82.00  & 52.00  & 72.00  & \textbf{70.00}  & \textbf{81.50}  \\
    \bottomrule
  \end{tabular}}%
  \label{table:ablation_ruler}
\end{table}

\textbf{Grouping} is the dominant factor on RULER, boosting the average from 40.47 to 81.28 (+40.81). RULER's needle targets (numbers, UUIDs) are split into multiple tokens; without grouping, scattered recomputation fragments these entities.

\begin{table}[h]
 \caption{Ablation Study of CacheClip on LongBench (Qwen2.5-14B-Instruct, Recomputation Ratio = 20\%).}
  \centering
  \begin{tabular}{ccc@{\hskip 1.2em}cccc}
    \toprule
    \multicolumn{3}{c}{Components} & \multicolumn{4}{c}{LongBench Scores} \\
    \cmidrule(lr){1-3} \cmidrule(lr){4-7}
    Recompute & Shared Prefix & Grouping & multifieldqa\_zh & 2wikimqa & hotpotqa & Avg. \\
    \midrule
    \multicolumn{3}{c}{\textit{Direct Reuse (no recomputation)}} & 49.44 & 37.70 & 44.63 & 43.92 \\
    \midrule
    \checkmark &              &              & \textbf{62.76} & 37.84 & 47.61 & 49.40 \\
    \checkmark & \checkmark   &              & 61.06 & \textbf{42.70} & 48.96 & 50.91 \\
    \checkmark &              & \checkmark   & 60.19 & 40.90 & 47.45 & 49.51 \\
    \checkmark & \checkmark   & \checkmark   & 62.43 & 41.40 & \textbf{52.12} & \textbf{51.98} \\
    \bottomrule
  \end{tabular}
  \label{table:ablation_longbench}
\end{table}

\textbf{Shared prefix} is the dominant factor on LongBench (\autoref{table:ablation_longbench}), yielding a +1.51 average gain (49.40$\to$50.91), concentrated on multi-hop QA (2wikimqa +4.86, hotpotqa +1.35). Grouping alone provides only marginal improvement (+0.11). LongBench's reasoning tasks are dominated by the attention sink problem: independently processed chunks develop strong attention to initial tokens, and concatenation introduces redundant sinks. Shared prefix resolves this by providing unified initial tokens, freeing attention capacity for cross-chunk reasoning.

The two components address complementary failure modes: local continuity loss (grouping) and global attention distortion (shared prefix). The full system achieves the best overall performance on both benchmarks.

\subsection{Prefill Efficiency Evaluation}

We benchmark CacheClip on a single NVIDIA L20 GPU with 16K input tokens (Qwen2.5-14B-Instruct, auxiliary: SmolLM2-135M-Instruct on Intel Xeon 6554S CPU).

\begin{table}[h]
 \caption{Prefill efficiency breakdown (input length = 16K, recomp\% = 20\%).}
  \centering
  \begin{tabular}{llllll}
    \toprule
    Method & Token Selection & Recomputation & Others & Total Time & Speedup  \\
    \midrule
    Full Attention & N/A & N/A & N/A & 5.641s & 1.00$\times$ \\
    CacheClip & 0.238s & 1.332s & 0.125s & 1.695s & 3.33$\times$ \\
    \bottomrule
  \end{tabular}
  \label{table:efficiency_evaluation}
\end{table}

CacheClip achieves 3.33$\times$ speedup (\autoref{table:efficiency_evaluation}). Token selection (0.238s) runs on CPU with precomputed auxiliary KV caches and is overlappable with GPU KV cache loading. Recomputation (1.332s, 78.6\% of total latency) uses a custom Triton~\cite{tillet2019triton} Flash Attention kernel supporting sparse query indices, as standard implementations assume dense contiguous queries. The recomputation ratio can be dynamically adjusted at runtime to trade off between quality and latency.

\section{Related Work}
\label{sec:related_work}

\paragraph{Prefix Caching and KV Cache Management.}
vLLM~\cite{kwon2023efficient} and SGLang~\cite{zheng2023efficiently} implement prefix caching at the serving layer, reusing KV caches when queries share identical prefixes. RAGCache~\cite{jin2024ragcache} extends this with a prefix tree over chunk sequences to maximize reuse in RAG pipelines. PromptCache~\cite{gim2024prompt} precomputes KV caches for individual text segments and concatenates them at inference time. While effective for static or repetitive prompts, these methods offer limited benefit in RAG scenarios where retrieved chunks and their orderings vary across queries (Section~\ref{subsec:kv_reuse_limits}).

\paragraph{KV Cache Calibration.}
APE~\cite{yang2025ape} and Zhang et al.~\cite{zhang2024attention} mitigate the repeated attention sink problem in concatenated KV caches by prepending a shared prefix to each chunk and retaining only one copy. APE further refines attention distributions by adjusting the attention temperature and scaling factor to sharpen focus on important tokens. Zhang et al.~\cite{zhang2024attention} additionally score chunk-level importance and discard the KV caches of less relevant chunks, trading potential information loss for sharper focus. These calibration methods improve attention consistency but cannot recover the missing inter-chunk attention needed for cross-document reasoning.

\paragraph{Selective Recomputation.}
CacheBlend~\cite{yao2024cacheblend} selectively recomputes a subset of tokens by identifying important positions through the primary model's early-layer signals, then blends the updated KV entries with precomputed ones. Cache-Craft~\cite{agarwal2025cache} stores multiple versions of each chunk's KV cache corresponding to different preceding contexts, selects the version with the most similar prefix, and recomputes tokens with the strongest inter-chunk connections. However, Cache-Craft's multi-version storage grows combinatorially with the number of chunks, making it impractical for large-scale RAG deployments. CacheClip differs from these methods by using an external auxiliary model for token selection rather than the primary model's own signals, and by grouping selected tokens into sliding windows to preserve local coherence during recomputation.

\paragraph{Finetuning for Local Attention.}
Block Attention~\cite{sun2024block} and TurboRAG~\cite{lu2024turborag} finetune LLMs to adapt to block-diagonal attention patterns, enabling independent chunk processing without quality loss. KVLink~\cite{yang2025kvlink} finetunes the model to bridge independently computed KV caches through learned linking tokens. While effective, these approaches require substantial training resources and task-specific datasets, limiting their generalizability when chunk configurations or task distributions change. CacheClip avoids finetuning the primary model entirely, relying on inference-time strategies for broad applicability.

\paragraph{Token Importance Identification.}
Several methods identify important tokens for KV cache compression during decoding or full-context prefilling. StreamingLLM~\cite{xiao2023efficient} retains initial and recent tokens to maintain generation quality in long sequences. H2O~\cite{zhang2023h2o} dynamically keeps tokens with the highest cumulative attention scores. SnapKV~\cite{li2024snapkv} uses attention from the last prompt segment to select top-$k$ relevant positions for compression. Quest~\cite{tang2024quest} estimates page-level token importance via key-value range tracking. PyramidKV~\cite{cai2024pyramidkv} applies layer-wise budgets based on the observation that attention sharpens in deeper layers. However, these methods assume access to the full attention context, making them inapplicable to RAG prefilling where inter-chunk attention is unavailable. CacheClip addresses this gap by using a lightweight auxiliary model's last-layer attention as a proxy for the primary model's deep-layer token importance.

\section{Conclusion and Future Work}
We present CacheClip, a practical framework that accelerates RAG inference while maintaining high generation quality. By leveraging auxiliary LLMs to identify critical tokens for selective KV cache recomputation, CacheClip achieves up to 3.33× speedup in prefill time with acceptable quality trade-offs, running auxiliary LLMs on head-node CPUs to avoid additional GPU overhead. As future work, fine-tuning the auxiliary model for attention pattern prediction could further improve token selection quality; since auxiliary models are typically under 1B parameters, the training cost is far lower than fine-tuning the primary LLM.

\bibliographystyle{unsrt}  
\bibliography{references}  

\newpage
\begin{appendices}
\section{Detailed Evaluation Results}

\begin{table}[H]
 \caption{Performance of full attention on RULER dataset for Qwen2.5-14B-Instruct model}
  \centering
  \resizebox{\textwidth}{!}{%
  \begin{tabular}{l rrr rrr rrr}
    \toprule
    Input Length & single1 & single2 & single3 & multikey1 & multikey2 & multikey3 & multivalue & multiquery & Average\\
    \midrule
    4096  & 100 & 100 & 100 & 100   & 100 & 100  & 97.75 & 100   & 99.72 \\
    8192  & 100 & 100 & 100 & 99.5  & 100 & 99.5 & 96.25 & 100   & 99.41 \\
    16384 & 100 & 100 & 100 & 99.5  & 100 & 99   & 92.75 & 99.88 & 98.89 \\
    \bottomrule
  \end{tabular}}%
  \label{tab:full_attn_ruler}
\end{table}


\begin{table}[H]
 \caption{Performance of APE on RULER dataset for Qwen2.5-14B-Instruct model}
  \centering
  \resizebox{\textwidth}{!}{%
  \begin{tabular}{l rrr rrr rrr}
    \toprule
    Input Length & single1 & single2 & single3 & multikey1 & multikey2 & multikey3 & multivalue & multiquery & Average\\
    \midrule
    4096  & 100 & 100 & 100 & 84 & 86 & 85 & 50   & 61.25 & 83.28 \\
    8192  & 100 & 100 & 100 & 73 & 65 & 57 & 34   & 46    & 71.88 \\
    16384 & 99  & 88  & 99  & 57 & 27 & 37 & 27.5 & 36    & 58.81 \\
    \bottomrule
  \end{tabular}}%
  \label{tab:ape_ruler}
\end{table}

\begin{table}[H]
 \caption{Performance of CacheBlend on RULER dataset for Qwen2.5-14B-Instruct model}
  \centering
  \resizebox{\textwidth}{!}{%
  \begin{tabular}{ll rrr rrr rrr}
    \toprule
    Recomp.\% & Input Length & single1 & single2 & single3 & multikey1 & multikey2 & multikey3 & multivalue & multiquery & Avg.\\
\midrule
10\% & 4096 & 100.00 & 100.00 & 89.69  & 92.00  & 87.00  & 72.00  & 60.94 & 88.64  & 86.28 \\
20\% & 4096 & 100.00 & 75.00  & 60.82  & 88.00  & 80.00  & 72.00  & 69.01 & 90.91  & 79.47 \\
30\% & 4096 & 100.00 & 31.25  & 28.87  & 71.00  & 84.00  & 75.00  & 66.93 & 91.92  & 68.62 \\
40\% & 4096 & 100.00 & 31.25  & 24.74  & 70.00  & 85.00  & 79.00  & 74.48 & 91.92  & 69.55 \\
50\% & 4096 & 100.00 & 31.25  & 24.74  & 71.00  & 92.00  & 86.00  & 79.69 & 93.43  & 72.26 \\
60\% & 4096 & 100.00 & 31.25  & 24.74  & 72.00  & 95.00  & 87.00  & 85.94 & 94.19  & 73.76 \\
70\% & 4096 & 100.00 & 31.25  & 24.74  & 72.00  & 97.00  & 87.00  & 89.58 & 93.94  & 74.44 \\
80\% & 4096 & 100.00 & 100.00 & 100.00 & 100.00 & 100.00 & 99.00  & 99.22 & 100.00 & 99.78 \\
90\% & 4096 & 100.00 & 100.00 & 100.00 & 100.00 & 100.00 & 99.00  & 99.22 & 99.75  & 99.75 \\
\midrule
10\% & 8192 & 98.00  & 98.98  & 89.58  & 84.00  & 73.00  & 57.00  & 40.89 & 78.00 & 77.43 \\
20\% & 8192 & 98.00  & 78.57  & 65.62  & 61.00  & 67.00  & 62.00  & 42.97 & 73.50 & 68.58 \\
30\% & 8192 & 99.00  & 15.31  & 23.96  & 33.00  & 68.00  & 62.00  & 47.66 & 74.50 & 52.93 \\
40\% & 8192 & 100.00 & 11.22  & 12.50  & 35.00  & 74.00  & 67.00  & 54.95 & 77.00 & 53.96 \\
50\% & 8192 & 100.00 & 11.22  & 12.50  & 37.00  & 84.00  & 68.00  & 61.46 & 80.00 & 56.77 \\
60\% & 8192 & 100.00 & 11.22  & 12.50  & 37.00  & 85.00  & 72.00  & 68.75 & 81.25 & 58.46 \\
70\% & 8192 & 100.00 & 11.22  & 12.50  & 44.00  & 90.00  & 77.00  & 88.28 & 98.75 & 65.22 \\
80\% & 8192 & 100.00 & 11.22  & 12.50  & 52.00  & 94.00  & 86.00  & 93.75 & 99.00 & 68.56 \\
90\% & 8192 & 100.00 & 100.00 & 100.00 & 100.00 & 100.00 & 100.00 & 97.92 & 99.75 & 99.71 \\
\midrule
20\% & 16384 & 98.00 & 83.00 & 72.00 & 54.00 & 30.00 & 35.00 & 23.50 & 72.00 & 58.44 \\
  \bottomrule
  \end{tabular}}%
  \label{tab:cacheblend_ruler}
\end{table}

\begin{table}[H]
 \caption{Performance of CacheClip on RULER dataset for Qwen2.5-14B-Instruct model (auxiliary model: SmolLM2-135M-Instruct)}
  \centering
  \resizebox{\textwidth}{!}{%
  \begin{tabular}{ll rrr rrr rrr}
    \toprule
    Recomp.\% & Input Length & single1 & single2 & single3 & multikey1 & multikey2 & multikey3 & multivalue & multiquery & Avg.\\
\midrule
10\% & 4096 & 79.00  & 100.00 & 99.00  & 98.00  & 97.00 & 81.00 & 59.25 & 77.75 & 86.38 \\
20\% & 4096 & 96.00  & 96.00  & 90.00  & 88.00  & 93.00 & 66.00 & 72.00 & 82.75 & 85.47 \\
30\% & 4096 & 99.00  & 88.00  & 84.00  & 87.00  & 86.00 & 58.00 & 79.00 & 78.75 & 82.47 \\
40\% & 4096 & 100.00 & 89.00  & 77.00  & 85.00  & 80.00 & 52.00 & 83.25 & 82.00 & 81.03 \\
50\% & 4096 & 100.00 & 95.00  & 75.00  & 89.00  & 71.00 & 58.00 & 86.00 & 81.75 & 81.97 \\
60\% & 4096 & 100.00 & 97.00  & 70.00  & 88.00  & 63.00 & 59.00 & 87.25 & 85.00 & 81.16 \\
70\% & 4096 & 100.00 & 98.00  & 67.00  & 93.00  & 61.00 & 67.00 & 89.25 & 90.75 & 83.25 \\
80\% & 4096 & 100.00 & 100.00 & 62.00  & 96.00  & 71.00 & 69.00 & 93.50 & 94.00 & 85.69 \\
90\% & 4096 & 100.00 & 100.00 & 61.00  & 100.00 & 88.00 & 81.00 & 95.25 & 95.00 & 90.03 \\
\midrule
10\% & 8192 & 77.00  & 99.00  & 100.00 & 96.00 & 88.00 & 72.00 & 47.75 & 64.50 & 80.53 \\
20\% & 8192 & 98.00  & 93.00  & 92.00  & 93.00 & 82.00 & 52.00 & 72.00 & 70.00 & 81.50  \\
30\% & 8192 & 99.00  & 93.00  & 80.00  & 90.00 & 76.00 & 43.00 & 74.00 & 73.75 & 78.59 \\
40\% & 8192 & 100.00 & 84.00  & 81.00  & 87.00 & 66.00 & 32.00 & 76.75 & 77.00 & 75.47 \\
50\% & 8192 & 100.00 & 89.00  & 74.00  & 92.00 & 65.00 & 45.00 & 81.75 & 78.75 & 78.19 \\
60\% & 8192 & 100.00 & 96.00  & 66.00  & 91.00 & 62.00 & 47.00 & 86.75 & 84.50 & 79.16 \\
70\% & 8192 & 100.00 & 99.00  & 65.00  & 97.00 & 59.00 & 54.00 & 90.50 & 88.50 & 81.62 \\
80\% & 8192 & 100.00 & 99.00  & 65.00  & 96.00 & 72.00 & 71.00 & 94.25 & 94.25 & 86.44 \\
90\% & 8192 & 100.00 & 100.00 & 61.00  & 99.00 & 90.00 & 84.00 & 96.50 & 97.50 & 91.00  \\
\midrule
10\% & 16384 & 65.00  & 88.00  & 90.00 & 84.00  & 79.00 & 57.00 & 34.75 & 62.50 & 70.03 \\
20\% & 16384 & 90.00  & 88.00  & 82.00 & 88.00  & 69.00 & 29.00 & 68.00 & 69.50 & 72.94 \\
30\% & 16384 & 97.00  & 83.00  & 71.00 & 84.00  & 64.00 & 26.00 & 72.75 & 72.75 & 71.31 \\
40\% & 16384 & 98.00  & 80.00  & 71.00 & 84.00  & 44.00 & 21.00 & 69.75 & 74.75 & 67.81 \\
50\% & 16384 & 98.00  & 91.00  & 67.00 & 86.00  & 51.00 & 25.00 & 75.00 & 83.75 & 72.09 \\
60\% & 16384 & 98.00  & 92.00  & 66.00 & 91.00  & 50.00 & 37.00 & 78.25 & 87.00 & 74.91 \\
70\% & 16384 & 99.00  & 97.00  & 72.00 & 97.00  & 49.00 & 47.00 & 86.75 & 92.75 & 80.06 \\
80\% & 16384 & 100.00 & 99.00  & 66.00 & 97.00  & 69.00 & 61.00 & 88.25 & 95.75 & 84.50  \\
90\% & 16384 & 100.00 & 100.00 & 67.00 & 100.00 & 75.00 & 76.00 & 90.00 & 98.25 & 88.28 \\
  \bottomrule
  \end{tabular}}%
  \label{tab:cacheclip_ruler_135M}
\end{table}

\begin{table}[H]
 \caption{Performance of CacheClip on RULER dataset for Qwen2.5-14B-Instruct model (auxiliary model: Qwen2.5-0.5B-Instruct)}
  \centering
  \resizebox{\textwidth}{!}{%
  \begin{tabular}{ll rrr rrr rrr}
    \toprule
    Recomp.\% & Input Length & single1 & single2 & single3 & multikey1 & multikey2 & multikey3 & multivalue & multiquery & Avg.\\
\midrule
10\% & 4096 & 100.00 & 100.00 & 93.00  & 95.00 & 99.00  & 88.00  & 74.00 & 85.75 & 91.84 \\
20\% & 4096 & 100.00 & 100.00 & 75.00  & 97.00 & 96.00  & 74.00  & 88.75 & 89.00 & 89.97 \\
30\% & 4096 & 100.00 & 100.00 & 69.00  & 94.00 & 92.00  & 72.00  & 96.00 & 94.50 & 89.69 \\
40\% & 4096 & 100.00 & 99.00  & 69.00  & 98.00 & 89.00  & 72.00  & 95.00 & 96.00 & 89.75 \\
50\% & 4096 & 100.00 & 100.00 & 65.00  & 97.00 & 87.00  & 73.00  & 97.75 & 97.50 & 89.66 \\
60\% & 4096 & 100.00 & 100.00 & 78.00  & 99.00 & 76.00  & 81.00  & 98.75 & 96.75 & 91.19 \\
70\% & 4096 & 100.00 & 100.00 & 80.00  & 98.00 & 83.00  & 90.00  & 98.75 & 97.00 & 93.34 \\
80\% & 4096 & 100.00 & 100.00 & 91.00  & 99.00 & 92.00  & 99.00  & 98.50 & 97.50 & 97.12 \\
90\% & 4096 & 100.00 & 100.00 & 100.00 & 99.00 & 99.00  & 100.00 & 98.00 & 99.00 & 99.38 \\
\midrule
10\% & 8192 & 100.00 & 99.00  & 89.00  & 94.00  & 89.00  & 80.00  & 74.00 & 77.25 & 87.78 \\
20\% & 8192 & 100.00 & 98.00  & 73.00  & 91.00  & 82.00  & 66.00  & 83.50 & 83.75 & 84.66 \\
30\% & 8192 & 100.00 & 98.00  & 73.00  & 98.00  & 79.00  & 55.00  & 93.75 & 91.75 & 86.06 \\
40\% & 8192 & 100.00 & 100.00 & 73.00  & 96.00  & 79.00  & 48.00  & 96.75 & 95.00 & 85.97 \\
50\% & 8192 & 100.00 & 99.00  & 72.00  & 99.00  & 79.00  & 52.00  & 98.25 & 98.00 & 87.16 \\
60\% & 8192 & 100.00 & 100.00 & 75.00  & 100.00 & 73.00  & 76.00  & 98.25 & 98.25 & 90.06 \\
70\% & 8192 & 100.00 & 100.00 & 90.00  & 100.00 & 86.00  & 89.00  & 98.75 & 98.25 & 95.25 \\
80\% & 8192 & 100.00 & 100.00 & 98.00  & 100.00 & 92.00  & 97.00  & 98.75 & 98.75 & 98.06 \\
90\% & 8192 & 100.00 & 100.00 & 100.00 & 100.00 & 100.00 & 100.00 & 98.00 & 99.00 & 99.62 \\
\midrule
10\% & 16384 & 90.00  & 94.00  & 86.00 & 86.00  & 73.00 & 61.00 & 66.50 & 70.75 & 78.41 \\
20\% & 16384 & 95.00  & 94.79  & 69.00 & 77.00  & 61.00 & 43.00 & 82.25 & 81.00 & 75.38 \\
30\% & 16384 & 95.00  & 96.00  & 68.00 & 90.00  & 61.00 & 42.00 & 89.25 & 86.25 & 78.44 \\
40\% & 16384 & 100.00 & 95.00  & 75.00 & 95.00  & 59.00 & 45.00 & 95.25 & 92.00 & 82.03 \\
50\% & 16384 & 100.00 & 99.00  & 78.00 & 98.00  & 65.00 & 59.00 & 94.50 & 93.25 & 85.84 \\
60\% & 16384 & 100.00 & 100.00 & 73.00 & 99.00  & 70.00 & 65.00 & 94.25 & 98.00 & 87.41 \\
70\% & 16384 & 100.00 & 100.00 & 82.00 & 100.00 & 80.00 & 83.00 & 94.25 & 98.75 & 92.25 \\
80\% & 16384 & 100.00 & 100.00 & 94.00 & 99.00  & 93.00 & 97.00 & 94.50 & 99.00 & 97.06 \\
90\% & 16384 & 100.00 & 100.00 & 99.00 & 100.00 & 94.00 & 99.00 & 94.50 & 98.00 & 98.06 \\
  \bottomrule
  \end{tabular}}%
  \label{tab:cacheclip_ruler_0.5B}
\end{table}

\begin{table}[H]
 \caption{Performance of CacheClip on RULER dataset for Qwen2.5-14B-Instruct model (sequence length 8192, auxiliary model: SmolLM2-135M-Instruct)}
  \centering
  \resizebox{\textwidth}{!}{%
  \begin{tabular}{lll rrr rrr rrr}
    \toprule
    Recomp.\% & Grouping & Shared Prefix & single1 & single2 & single3 & multikey1 & multikey2 & multikey3 & multivalue & multiquery & Avg. \\
\midrule
10\% & $\times$ & $\times$     & 40.00  & 35.00  & 55.00  & 46.00  & 85.00  & 52.00  & 29.75  & 41.25  & 48.00  \\
10\% & $\times$ & \checkmark   & 38.00  & 41.00  & 54.00  & 49.00  & 83.00  & 40.00  & 31.00  & 46.00  & 47.75  \\
10\% & \checkmark & $\times$   & 90.00  & 99.00  & 88.00  & 90.00  & 88.00  & 75.00  & 48.25  & 65.25  & 80.44  \\
10\% & \checkmark & \checkmark & 77.00  & 99.00  & 100.00 & 96.00  & 88.00  & 72.00  & 47.75  & 64.50  & 80.53  \\
\midrule
20\% & $\times$ & $\times$     & 78.00  & 20.00  & 39.00  & 33.00  & 77.00  & 20.00  & 25.25  & 31.50  & 40.47  \\
20\% & $\times$ & \checkmark   & 64.00  & 20.00  & 46.00  & 31.00  & 74.00  & 16.00  & 26.25  & 32.00  & 38.66  \\
20\% & \checkmark & $\times$   & 98.00  & 92.00  & 81.00  & 94.00  & 86.00  & 55.00  & 74.50  & 69.75  & 81.28  \\
20\% & \checkmark & \checkmark & 98.00  & 93.00  & 92.00  & 93.00  & 82.00  & 52.00  & 72.00  & 70.00  & 81.50  \\
\midrule
30\% & $\times$ & $\times$     & 96.00  & 26.00  & 46.00  & 27.00  & 69.00  & 16.00  & 24.00  & 29.25  & 41.66  \\
30\% & $\times$ & \checkmark   & 88.00  & 23.00  & 56.00  & 27.00  & 64.00  & 15.00  & 23.00  & 26.00  & 40.25  \\
30\% & \checkmark & $\times$   & 98.00  & 90.00  & 78.00  & 90.00  & 76.00  & 45.00  & 75.00  & 73.25  & 78.16  \\
30\% & \checkmark & \checkmark & 99.00  & 93.00  & 80.00  & 90.00  & 76.00  & 43.00  & 74.00  & 73.75  & 78.59  \\
\midrule
40\% & $\times$ & $\times$     & 99.00  & 36.00  & 45.00  & 35.00  & 63.00  & 15.00  & 28.50  & 31.75  & 44.16  \\
40\% & $\times$ & \checkmark   & 98.00  & 31.00  & 58.00  & 38.00  & 56.00  & 15.00  & 27.50  & 32.75  & 44.53  \\
40\% & \checkmark & $\times$   & 99.00  & 89.00  & 75.00  & 89.00  & 66.00  & 36.00  & 80.25  & 78.25  & 76.56  \\
40\% & \checkmark & \checkmark & 100.00 & 84.00  & 81.00  & 87.00  & 66.00  & 32.00  & 76.75  & 77.00  & 75.47  \\
\midrule
50\% & $\times$ & $\times$     & 99.00  & 46.00  & 43.00  & 55.00  & 57.00  & 19.00  & 35.50  & 38.50  & 49.12  \\
50\% & $\times$ & \checkmark   & 99.00  & 45.00  & 52.00  & 43.00  & 48.00  & 21.00  & 36.50  & 37.50  & 47.75  \\
50\% & \checkmark & $\times$   & 99.00  & 93.00  & 70.00  & 90.00  & 68.00  & 40.00  & 82.75  & 77.25  & 77.50  \\
50\% & \checkmark & \checkmark & 100.00 & 89.00  & 74.00  & 92.00  & 65.00  & 45.00  & 81.75  & 78.75  & 78.19  \\
\midrule
60\% & $\times$ & $\times$     & 100.00 & 65.00  & 39.00  & 58.00  & 55.00  & 21.00  & 42.75  & 44.50  & 53.16  \\
60\% & $\times$ & \checkmark   & 99.00  & 58.00  & 48.00  & 56.00  & 45.00  & 23.00  & 45.50  & 45.75  & 52.53  \\
60\% & \checkmark & $\times$   & 100.00 & 98.00  & 69.00  & 93.00  & 66.00  & 46.00  & 86.75  & 85.50  & 80.53  \\
60\% & \checkmark & \checkmark & 100.00 & 96.00  & 66.00  & 91.00  & 62.00  & 47.00  & 86.75  & 84.50  & 79.16  \\
\midrule
70\% & $\times$ & $\times$     & 100.00 & 76.00  & 34.00  & 77.00  & 44.00  & 33.00  & 51.00  & 54.50  & 58.69  \\
70\% & $\times$ & \checkmark   & 99.00  & 73.00  & 36.00  & 74.00  & 43.00  & 29.00  & 55.50  & 57.50  & 58.38  \\
70\% & \checkmark & $\times$   & 100.00 & 99.00  & 64.00  & 95.00  & 65.00  & 58.00  & 89.75  & 91.50  & 82.78  \\
70\% & \checkmark & \checkmark & 100.00 & 99.00  & 65.00  & 97.00  & 59.00  & 54.00  & 90.50  & 88.50  & 81.62  \\
\midrule
80\% & $\times$ & $\times$     & 100.00 & 89.00  & 28.00  & 90.00  & 47.00  & 44.00  & 66.50  & 71.00  & 66.94  \\
80\% & $\times$ & \checkmark   & 100.00 & 89.00  & 27.00  & 83.00  & 42.00  & 45.00  & 67.75  & 71.00  & 65.59  \\
80\% & \checkmark & $\times$   & 100.00 & 100.00 & 62.00  & 97.00  & 81.00  & 70.00  & 92.75  & 93.75  & 87.06  \\
80\% & \checkmark & \checkmark & 100.00 & 99.00  & 65.00  & 96.00  & 72.00  & 71.00  & 94.25  & 94.25  & 86.44  \\
\midrule
90\% & $\times$ & $\times$     & 100.00 & 93.00  & 26.00  & 94.00  & 66.00  & 65.00  & 80.75  & 84.50  & 76.16  \\
90\% & $\times$ & \checkmark   & 100.00 & 93.00  & 28.00  & 93.00  & 69.00  & 65.00  & 83.75  & 85.50  & 77.16  \\
90\% & \checkmark & $\times$   & 100.00 & 100.00 & 58.00  & 100.00 & 92.00  & 83.00  & 94.75  & 96.00  & 90.47  \\
90\% & \checkmark & \checkmark & 100.00 & 100.00 & 61.00  & 99.00  & 90.00  & 84.00  & 96.50  & 97.50  & 91.00  \\
  \bottomrule
  \end{tabular}}%
  \label{tab:cacheclip_ruler_ab}
\end{table}

\begin{table}[H]
 \caption{Performance of Methods on LongBench dataset for Qwen2.5-14B-Instruct model}
  \centering
  \begin{tabular}{llll}
    \toprule
    Method & multifieldqa\_zh & 2wikimqa & hotpotqa\\
    \midrule
   Direct Reuse & 49.44 & 37.70 & 44.63 \\                          
   APE & 59.70 & 38.34 & 45.29 \\ 
   CacheBlend (recomp\%=20\%) & 57.34 & 41.08 & 44.11 \\
   CacheClip (recomp \%=20\%, auxiliary model: SmolLM2-135M-Instruct) & 62.43 & 41.40 & 52.12 \\
   CacheClip (recomp \%=20\%, auxiliary model: Qwen2.5-0.5B-Instruct) & 61.37 & 46.04 & 55.74 \\
   Full Attention & 64.93 & 54.36 & 59.71 \\
    \bottomrule
  \end{tabular}
  \label{tab:longbench}
\end{table}

\section{1D Attention Distribution Plots}
\label{sec:attention_1d_curve}

See \autoref{fig:attention_1d_curve} in Section~\ref{subsec:small_but_aligned} for the 1D attention distribution plots. The auxiliary model's last layer and the primary model's last layer show highly correlated peaks (Pearson $r = 0.97$, Spearman $\rho = 0.60$), while the primary model's own first layer is less aligned with its last layer (Pearson $r = 0.70$, Spearman $\rho = 0.54$).

\section{KL Divergence Analysis}
\label{sec:kl_results}

See \autoref{tab:kl_results} in Section~\ref{subsec:small_but_aligned} for the full KL divergence analysis. KL confirms distributional alignment for same-primary pairs (A, B); cross-family pairs show higher KL due to architectural style differences, which does not affect position-level agreement (see Jaccard in \autoref{tab:jaccard_results}).

\end{appendices}

\end{document}